\documentclass[runningheads]{llncs}
\usepackage{graphicx}
\usepackage{xcolor}
\usepackage{array}
\usepackage{pgfplots}
\usetikzlibrary{patterns}
\pgfplotsset{width=7cm,compat=1.8}
\newcolumntype{P}[1]{>{\centering\arraybackslash}p{#1}}
\begin{document}
\title{Walk in Wild: An Ensemble Approach for Hostility Detection in Hindi Posts}
%
%\titlerunning{Abbreviated paper title}
% If the paper title is too long for the running head, you can set
% an abbreviated paper title here
%
\author{Chander Shekhar \and
Bhavya Bagla\and
Kaushal Kumar Maurya\and \\
Maunendra Sankar Desarkar }
\authorrunning{Chander et al.}
% First names are abbreviated in the running head.
% If there are more than two authors, 'et al.' is used.
%
\institute{Indian Institute of Technology Hyderabad, India \\
\email{cs17btech11011@iith.ac.in}, \email{cs17btech11007@iith.ac.in}, \email{cs18resch11003@iith.ac.in}, \email{maunendra@cse.iith.ac.in}
}
\maketitle              % typeset the header of the contribution
\begin{abstract}
As the reach of the internet increases, pejorative terms started flooding over social media platforms. This leads to the necessity of identifying hostile content on social media platforms. Identification of hostile contents on low-resource languages like Hindi poses different challenges due to its diverse syntactic structure compared to English. In this paper, we develop a simple ensemble based model on pre-trained mBERT and popular classification algorithms like Artificial Neural Network (ANN) and XGBoost for hostility detection in Hindi posts.  We formulated this problem as \textit{binary} classification (hostile and non-hostile class) and \textit{multi-label multi-class} classification problem (for more fine-grained hostile classes). We received third overall rank in the competition\footnote{\url{https://competitions.codalab.org/competitions/26654}} and weighted F1-scores of $\sim{}$0.969 and $\sim{}$0.61 on the binary and  multi-label multi-class classification tasks respectively.

\end{abstract}

\section{Introduction}
During coronavirus lockdown, number of active internet users across the globe has increased rapidly. The government enforced lockdown, which pushed people to stay indoors and thus, increased the engagement with social media platforms like Facebook, Twitter, Instagram, Whatsapp, etc. This led to increased hostile posts over social media, including cyberbullying, trolling, spreading hate, death threat, etc. A major challenge for the common users in the digital space is to identify misinformation (aka fake news) in online content. In addition to that, according to a recent survey\footnote{\url{https://bit.ly/38BBrTu}}, on Twitter, there has been a 900\% increase in hate speech directed towards Chinese people and 200\% increase in the traffic to Hate sites and posts written against the Asian community. It is also found\footnote{\url{http://bit.ly/38Eo7gY}} that the percentage of non-English tweets in India has jumped up by 50\%. This inspires a necessity of research in hostility detection in posts written in low resource but widely-used languages like Hindi.

As billions of posts appear each day on social media and anti-social elements can get full anonymity while expressing hostile behavior over the internet, identification of authorized information should have a reliable automation system. Even though Hindi is the third most spoken language globally, it is considered a low-resource language due to the unavailability of accurate tools and suitable datasets for various tasks in Hindi. This motivates us to take the task of hostility detection of Hindi posts on Social media.  %We found that there are not many datasets published for hostility detection in Hindi. Moreover, the published ones(Jha et al. (2020)\cite{JHA20202324},(Safi Samghabadi et al. (2020)\cite{safi_2020}) have either of the following problems: scarcity of data, hostility detection is focused only on a specific sub-domain.
%Along with this, as Hindi is written in Devanagari script, there are not enough modules to be used for pre-processing and embedding generation, so to cut it short, the options are limited, or one has to work on it from scratch.

We view the hostility detection as a two-stage process. First, a Coarse-grained classification is done to mark a post as hostile or non-hostile. If the post is detected as hostile, then the second stage performs a more fine-grained classification of hostile classes. We briefly define these two terms: %Our task is to perform multi-label multi-class classification on textual data published by Bhardwaj et al. (2020) \cite{bhardwaj2020hostility}. 
 %and non-hostile. Hostile posts can belong to one or more of these hostile classes, although non-hostile posts cannot belong to any hostile classes. Thus, the two subtasks derived from the problem are as follows:
\begin{enumerate}
    \item \textbf{Coarse-grained classification}: It is a binary classification problem in which each post is categorised as hostile or non-hostile.
    \item \textbf{Fine-grained classification}: It is a multi-label multi-class classification of the hostile classes. Each hostile post belongs to one or more of the following categories:  fake news, hate speech, offensive and defamation.
\end{enumerate}

In our proposed approach, we leverage the pre-trained multilingual BERT (mBERT) \cite{devlin2019bert}\footnote{\url{https://huggingface.co/bert-base-multilingual-uncased}} for input post representation and further these representations are used as input for Artificial Neural Network (aka ANN) and other ML learning models for binary and multi-label multi-class classification problems. The base architecture of mBERT is the same as BERT. BERT and mBERT have proven as the state of the art models across multiple NLU and NLG tasks.  %A question arises, Why m-BERT? The answer to this question is that there are state-of-the-art models like word2vec and GloVe, but they fail to grab contextual information. Whereas, BERT is bi-directionally trained and can capture the context adequately.  Consider this, for instance, Hindi words like {\it chhakka} translates to a group of six, but at the same time, it is considered to be ultra-offensive towards the LQBTQ community.

The rest of this paper is organized in the following way: Section 2 presents related work; Section 3 gives an in-depth explanation of our model. Section 4 presents the experimental setup. Section 5 provides results and our analysis. Finally, we provide conclusions and directions for future work in Section 6.

\section{Related Work}
A hostility detection dataset in Hindi was presented in  \cite{bhardwaj2020hostility}. The authors develop their hostility detection models using traditional machine learning algorithms, i.e., Support Vector Machine (SVM), Decision Tree (DT), Random Forest (RF), and Logistic Regression (LR), on top of the post embeddings extracted using pre-trained m-BERT. Multiple profound researchers have tried to tackle this problem of hostility detection in the past few years \cite{waseem-hovy-2016-hateful}, \cite{davidson2017automated}, etc. However, their research was limited to the English Language. Some aspects of detecting hate speech that was limited to racism and sexism using character n-gram were discussed in \cite{waseem-hovy-2016-hateful}. In \cite{davidson2017automated}, the authors use crowd-sourcing to collect lexicon for hateful/offensive language and annotate the tweets as hate, offensive or none. Their model was based on keywords. They were able to detect hateful/offensive language if the text contains hate words explicitly. There has been several recent advancements for hostility/toxicity detection in non-English languages, such as Brazilian Portuguese \cite{leite2020toxic}, Hindi \cite{JHA20202324}, Bengali \cite{hossain-etal-2020-banfakenews}), etc. These works focus on generating new annotated dataset in a particular language. They also provide benchmarks for the datasets to facilitate further research in these languages.
 
\section{Methodology}
\begin{figure}[h!]
  \centering
  \includegraphics[width=\linewidth]{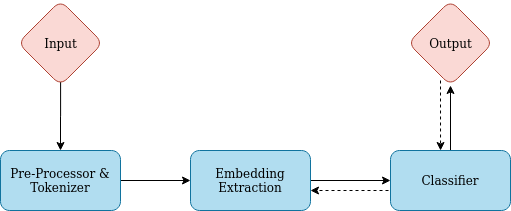}
  \caption{The flow diagram for coarse-grained and fine-grained classification task}
  \label{fig:flow_digm}
\end{figure}

A flow diagram for our coarse-grained and fine-grained classification techniques is included in figure-\ref{fig:flow_digm}. It consists of three stages: Pre-processing, embeddings extraction, and classification model. 

\subsection{Pre-Processing}
In general, tweets and online posts contain unstructured language/sentence, in which each token consists of meaningful information. Removal of any token may distort the meaning of the sentence. The dataset consists of tokens like emojis, hashtags, URLs, etc. Experimental results show that emojis and hashtags are essential features for hostility detection, whereas URLs are not. We manually removed all the URLs in the pre-processing step. The pre-processed posts are tokenized using the \textit{mBERT-tokenizer}. We choose input post length threshold as 128 and truncate (or pad) the post if it is longer (or shorter). %To make all the tokenized sequences of equal length, we have to choose a threshold. If a sequence has a smaller length than the threshold, then they are either padded with zeros. Else if their length is greater than the threshold, they are truncated. We choose this threshold to be 128 based on the average length of tokenized sequences and experimental results.

\subsection{Post Embedding Extraction }
We extracted Post embedding in two ways:
\begin{enumerate}
    \item \textbf{Raw Representation:} The pre-processed and tokenized input post was fed through the mBERT\cite{devlin2019bert} model. We obtained $[CLS]$ representation from mBERT as final post representation of dimension 768.
    \item \textbf{Fine-tuned Representation: } The extracted \textit{Raw Representation} of each post were used to train ANN model for both the classification tasks. Once the model is trained we performed single forward pass of Raw Representations to obtain Fine-tuned Representation.
\end{enumerate}
  %For each post  The dictionary contains \(last\_hidden\_state\) and \(pooler\_output\). \(last\_hidden\_state\) contains the contextual embeddings of every token in the sequence. \(pooler\_output\) contains the embeddings of the sequence's first token(classification token), further processed by a Linear layer and a Tanh activation function. Classification token's embeddings, as the name suggests, are pre-trained for classification. Hence, we use them for further classification and omit the word-level embeddings. Now, each post is a 768-dimensional vector.

\subsection{Classification Models}\label{subsec:cls_model}
\begin{figure*}[h!]
    \centering
    \includegraphics[width=\textwidth]{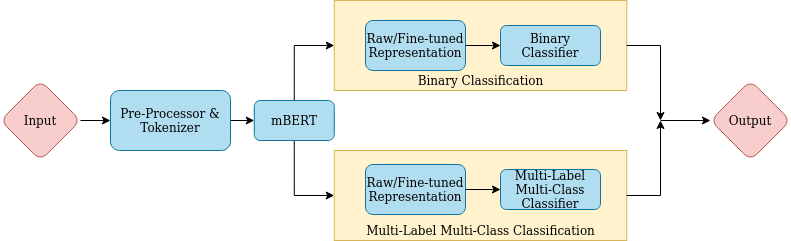}
    \caption{Architectural  Diagram of proposed model}
    \label{fig:arc_digm}
\end{figure*}

An architectural diagram of proposed model is presented in figure-\ref{fig:arc_digm}.

\subsubsection{ANN and XGBoost based Models:} The \textit{Raw Representation} of each post is an input for the Artificial Neural Network (ANN) classifier. For Coarse-grained classification, we applied \textit{softmax} layer on top of ANN representation to obtain hostile and non-hostile class probabilities. In Fine-grained tasks, a \textit{sigmod} layer with four neurons is used on top of ANN to obtain probability of four hostile classes. In another setting, we extracted the \textit{Fine-tuned Representation} from ANN, which acts as an input for XGboost classifier. We trained XGboost classifier for both binary and multi-label class classification task to obtain final label distributions. 

%In Coarse-grained, we pass the input representation from linear layer  fallowed by softmax function to obtain probabilities of each hostile and non-hostile classes.  As Fine-grained is multi-label multi-class classification problem, to handle this we feed input post representation through XGBoost classifier and used sigmod with four neurons to obtain probabilities of each hostile classes. \textcolor{red}{it is unclear how this has been done, please add more details}  
%first and then  model's weights according to the dataset. The training process is entirely different for a DNN model like BERT and XGBoost classifier. So, these models cannot be trained together. We use the m-BERT model finetuned alongside the training of ANN classifier and combine it with the XGBoost classifier. This combination allows us to leverage the m-BERT model's finetuning while generating embeddings for the XGBoost classifier.
%\subsubsection{XGboost Based Models}

\subsubsection{Ensemble Model: } \label{subsec: ensb}
%We take a weighted average of all the previous four submissions  results.  The  fine-grained  f1-score  on  the  validation  dataset  isused as the weights in the weighted average. The following formula is used for taking the average.

We propose a novel ensemble model to effectively combine existing model's output. Let's assume that there are $n$ posts in the test/validation dataset and $m$ existing models with outputs $<O_{11}, O_{12}, ...O_{1n}>$, $<O_{21}, O_{22}, ...O_{2n}>$, ..... $<O_{m1}, O_{m2}, ...O_{mn}>$. Where $O_{ij}$ is multi-hot vector output from $i^{th}$ model for $j^{th}$ post. The multi-hot vector consists of binary value for all classes (i.e., non-hostile, fake, hate, offensive and defamation). We also obtain fine-grained f1-scores from all $m$ models as $ff1_1$, $ff1_2$, $ff1_3$,... $ff1_m$. Note that $ff1_i$ is computed on validation dataset and same values were used for test data (as test data label are not available for $ff1_i$ computation). The ensembled multi-hot output $\hat{O_{ij}}$ is obtained in the following  way:
\begin{equation}
    \hat{O_{ij}} = \frac{\sum_{i=1}^{m} \sum_{j=1}^{n} O_{ij} \times ff1_i}{\sum_{i=1}^{m} ff1_i} 
\end{equation}

%According to the architecture given above, we train two models: The binary classification model and the multi-label classification model. For simplicity, we will refer to the Binary classification model as the Binary model and the Multi-label classification model as the multi-label model.

%We ensemble the two models (as shown in Figure 2) based on the following criteria:
%\begin{itemize}
%    \item If the binary model predicts non-hostile, then our model outputs non-hostile.
%    \item Our model outputs all the hostile classes predicted by the multi-label model if the binary model predicts hostility.
%\end{itemize}

%This ensembling method allows us to leverage the binary model's high prediction accuracy, while the multi-label model performs the fine-grained prediction. The main idea is that the binary model's hostility prediction supersedes the prediction of the multi-label model.

\section{Experimental Set-up}
\subsection{Data}
We are using a dataset released by  \cite{bhardwaj2020hostility}. The dataset is a collection of $\sim{}$8200 social media posts in Hindi. Each post belongs to either hostile (i.e., fake, defamation, hate and offensive) or non-hostile class, where each hostile post can have more than one hostile label. \cite{bhardwaj2020hostility} defines the classes as follows:
\begin{itemize}
\item \textbf{Fake News}: A claim or information that is verified to be not true. 
\item \textbf{Hate Speech}: A post that targets a specific group of people based on their race, geographical belonging,  ethnicity, religious beliefs, etc., with malicious intentions of spreading hatred or encouraging violence. 
\item \textbf{Offensive}: A post containing rude, profane, impolite, or vulgar language to insult a targeted group or individual.
\item \textbf{Defamation}: A post containing mis-information regarding an individual or group, which is made in an attempt to destroy their reputation publicly. 
\item \textbf{Non-Hostile}: A post that is not hostile.

\end{itemize}
Dataset statistics are included in Table-\ref{tab:data_stats}. For coarse-grained classification, the data is balanced as the number of examples for the two classes are almost similar. For fine-grained classification, the distribution of data points across all hostile classes is not uniform.

\begin{center}
    \begin{table}
    \centering
      \begin{tabular}{c|c|c|c|c|c|c}
        \hline\hline
         & \textbf{Fake} & \textbf{Hate} & \textbf{Offense} & \textbf{Defame} & \textbf{\#Hostile} & \textbf{\#Non-Hostile} \\
        \hline \hline
        Train & 1144 & 792 & 742 & 564 & 2678 & 3050 \\
        Validation & 160 & 103 & 110 & 77 & 376 & 435 \\
        Test & 334 & 237 & 219 & 169 & 780 & 873 \\
        \hline
        Overall & 1638 & 1132 & 1071 & 810 & 3834 & 4358 \\
        \hline \hline
      \end{tabular}
      \caption{Dataset Statistics}
      \label{tab:data_stats}
    \end{table}
\end{center}

\subsection{Evaluation Metrics}
The \textbf{Weighted F1-score} is our primary metric to evaluate the proposed models. For coarse-grained evaluation, weighted F1-score is computed for hostile and non-hostile classes, whereas for fine-grained evaluation, it is computed across four hostile classes. 
%\begin{itemize}
%    \item \textbf{Coarse-grained evaluation}: Weighted F1-score of hostile and non-hostile classes. A post belonging to any of the hostile classes is treated as hostile.
%    \item \textbf{Fine-grained evaluation:} Weighted F1-score of all the four hostile classes.
%\end{itemize}
\[ F1-Score = \frac{2(precision \times recall)}{(precision + recall)} \]
Weighted F1-Score is the weighted average of F1-Scores of all the classes by support (the number of true instances for each label). We also report individual F1-score for all hostile classes (see Tables \ref{tab:val_ruslt} and \ref{tab:test_ruslt}).

\subsection{Baseline}
%The results provided by the dataset authors\cite{bhardwaj2020hostility} are treated as the baseline to compare our models. Before training their model, they remove stopwords, non-alphanumeric characters, emojis, URLs as part of pre-processing. They use m-BERT to generate word embeddings for each word in the sentence. The sentence is represented as an average of its constituent word embeddings.
We obtained baseline models from \cite{bhardwaj2020hostility}. Similar to our approach, authors first extracted the post representation from mBERT and applied following algorithms: \textbf{Support Vector Machine (SVM)}, \textbf{Decision Tree (DT)}, \textbf{Random Forest (RF)}, and \textbf{Logistic Regression (LR)} for both the tasks. They reported weighted f1-score on the validation dataset.% All the algorithms are implemented using Scikit-Learn with the following hyper-parameters. For SVM, a linear kernel with \(c=0.01\) and \(\gamma = 1\). For Random Forests, the number of estimators is 400. For LogisticRegression, 'l1' penalty with 'liblinear' solver. In MLP, two hidden layers having 30 and 10 neurons each are used with a softmax layer. 'ReLU' is used as the activation function and a learning rate of 0.001. For each case, the rest of the parameters are default.

%\begin{center}
%    \begin{table}
%    \centering
%      \caption{Coarse-grained and Fine-grained weighted F1-score on the validation dataset}
%      \label{tab2}
%      \begin{tabular}{|P{1.5cm}|P{1.5cm}|P{1.6cm}|P{1.5cm}|P{1.5cm}|P{1.5cm}|P{1.5cm}|}
%        \hline
%        Model & Coarse Grained F1-Score & Defamation F1-Score & Fake F1-Score & Hate %F1-Score & Offensive F1-Score & Fine-Grained F1-Score \\
%%        \hline
 %       LR & 0.8398 & 0.4427 & \textbf{0.6815} & 0.3876 & 0.3627 &  0.4954\\
 %       SVM & \textbf{0.8411} & \textbf{0.4749} & 0.6644 & \textbf{0.4198} & \textbf{0.4357} & \textbf{0.5201} \\
 %       RF & 0.7979 & 0.0683 & 0.5343 & 0.0701 & 0.0256 & 0.2240 \\
 %       MLP & 0.8345 & 0.3482 & 0.6603 & 0.4069 & 0.2941 & 0.4594 \\
 %       \hline
 %     \end{tabular}
 %   \end{table}
%\end{center}

\subsection{Implementation Details}\label{subsec:impdet}
For ANN classifier, we use AdamW Optimizer with weight decay as 0.001, input size 128, learning rate as 0.00003, one hidden layer with 256 neurons, and the 'ReLU' activation function. After generating sentence embeddings, we apply a dropout to avoid overfitting.  Binary cross-entropy is used as the loss function with logits. For the XGBoost classifier, the learning objective is set as 'binary: logistic' and the max\_depth as 4. 

\section{Results and Analysis}

\subsection{Submission and Leader-board}
In the competition, we submitted our best five results which are variant of the proposed models (see section-\ref{subsec:cls_model}). The details of all five submissions are enumerated below:
\begin{enumerate}
    \item \textbf{Submission-1:} This is the basic ANN model with \textit{Raw Representation }. We trained this model for five epochs with dropout of 0.2. Rest of the hyper-parameters were same as mentioned in Section \ref{subsec:impdet}. 
    \item \textbf{Submission-2:} The model architecture was similar to Submission-1. The hyper-parameter setting was different. Unlike submission-1 we trained the model for 10 epochs with dropout of 0.3. There are no changes in other parameters. 
    \item \textbf{Submission-3:} In this model we applied XGboost model with \textit{Fine-tuned Representation}. Here we extracted the \textit{Fine-tuned Representation} for each post from Submission-1 model. Rest of the hyper-parameters were same as section \ref{subsec:impdet}.
    \item \textbf{Submission-4:} The model architecture of this submission was similar to submission-3. Unlike submission-3, we extracted the \textit{Fine-tuned Representation} for each post from submission-2 model. There were no changes in other parameters. 
    \item \textbf{Submission-5:} Our fifth submission approach is similar to ensemble model proposed in Section  \ref{subsec: ensb}. Only difference is that we took model outputs from above four submissions. The updated equation is:
    \begin{equation}
    \hat{O_{ij}} = \frac{\sum_{i=1}^{4} \sum_{j=1}^{n} O_{ij} \times ff1_i}{\sum_{i=1}^{4} ff1_i} 
    \end{equation}
\end{enumerate}

\begin{table}
  \caption{Results of our top 5 submissions on the validation dataset}
  \label{tab3}
  \scalebox{0.8}{
  \begin{tabular}{ |P{2.03cm}|P{2.03cm}|P{2.03cm}| P{2.03cm}|P{2.03cm}|P{2.03cm}| P{2.03cm}|}
    \hline
    Model/ Submission No. & Coarse Grained F1-Score & Defamation F1-Score & Fake F1-Score & Hate F1-Score & Offensive F1-Score & Weighted Fine-grained F1-Score \\
    \hline
    1 & 0.9692 & 0.2656 & 0.8254 & 0.4876 & 0.5326 & 0.58 \\
    2 & 0.9654 & 0.3459 & 0.8274 & \textbf{0.5388} & 0.5405 & 0.6088 \\
    3 & \textbf{0.9692} & 0.2483 & 0.8217 & 0.4896 & \textbf{0.5631} & 0.5832 \\
    4 & 0.9692 & \textbf{0.3810 }& 0.8365 & 0.5172 & 0.55 & \textbf{0.6149} \\
    5 & 0.9692 & 0.3247 & \textbf{0.8387} & 0.5214 & 0.5426 & 0.6054 \\
    \hline
  \end{tabular}\label{tab:val_ruslt}
  }
\end{table}

\begin{table}
  \caption{Results of our top 5 submissions on the test dataset}
  \label{tab4}
  \scalebox{0.8}{
  \begin{tabular}{ |P{2.03cm}|P{2.03cm}|P{2.03cm}| P{2.03cm}|P{2.03cm}|P{2.03cm}| P{2.03cm}|}
    \hline 
    Model/ Submission No. & Coarse Grained F1-Score & Defamation F1-Score & Fake F1-Score & Hate F1-Score & Offensive F1-Score & Weighted Fine-grained F1-Score \\
    \hline
    1 & 0.9612 & 0.3564 & 0.7823 & \textbf{0.5556} & 0.578 & 0.6047  \\
    2 & \textbf{0.9691} & 0.3061 & 0.7915 & 0.4282 & 0.5699 & 0.566 \\
    3 & 0.9655 & 0.3544 & \textbf{0.8} & 0.4129 & 0.5816 & 0.5764 \\
    4 & 0.9655 & \textbf{0.4343} & 0.7838 & 0.525 & 0.5661 & \textbf{0.6088}  \\
    5 & 0.9655 & 0.3765 & 0.7844 & 0.5339 & \textbf{0.5854} & 0.6054 \\
    \hline
  \end{tabular}\label{tab:test_ruslt}
  }
\end{table}

Table~\ref{tab3} and Table~\ref{tab4} summarize our results on the validation and test dataset, respectively submitted to the competition. We observed an almost similar trend of f1-scores across the test and validation dataset. We found that the best f1-scores for different labels are distributed across different submissions. Though, submission results across different submissions have minor changes. Coarse-grained classification received the highest score as this is only a binary classification set-up. On weighted Fine-grained F1-Score, submission-4 received the best score, which is close to the ensemble method.

\subsection{Comparisons with Baseline}
Our model consistently outperformed all the baseline models on validation dataset. The baseline results are not available for test dataset. The comparison results are included in Table-\ref{tab:comp}.  The defamation class poses lowest score mostly because it consists of minimal number of training points.  Except, defamation class, all our proposed models consistently outperformed all the baselines across all the categories by large margin. The f1-score is highest for the fake label in hostile categories because fake has highest number of training points (see Table-\ref{tab:data_stats}). We can infer that the the model performance is heavily dependent on the number of training points even with large multi-lingual pre-trained models like mBERT.   

\begin{table}
  \caption{Comparison of our top 5 submissions with the baseline models on the validation dataset}
  \label{tab6}
  \scalebox{0.8}{
  \begin{tabular}{ |P{2.03cm}|P{2.03cm}|P{2.03cm}| P{2.03cm}|P{2.03cm}|P{2.03cm}| P{2.03cm}|}
    \hline
    Model/ Submission No. & Coarse Grained F1-Score & Defamation F1-Score & Fake F1-Score & Hate F1-Score & Offensive F1-Score & Weighted Fine-grained F1-Score \\
    \hline
    LR & 0.8398 & 0.4427 & 0.6815 & 0.3876 & 0.3627 &  0.4954\\
    SVM & 0.8411 & \textbf{0.4749} & 0.6644 & 0.4198 & 0.4357 & 0.5201 \\
    RF & 0.7979 & 0.0683 & 0.5343 & 0.0701 & 0.0256 & 0.2240 \\
    MLP & 0.8345 & 0.3482 & 0.6603 & 0.4069 & 0.2941 & 0.4594 \\
    \hline
    1 & 0.9692 & 0.2656 & 0.8254 & 0.4876 & 0.5326 & 0.58 \\
    2 & 0.9654 & 0.3459 & 0.8274 & \textbf{0.5388} & 0.5405 & 0.6088 \\
    3 & \textbf{0.9692} & 0.2483 & 0.8217 & 0.4896 & \textbf{0.5631} & 0.5832 \\
    4 & 0.9692 & 0.3810 & 0.8365 & 0.5172 & 0.55 & \textbf{0.6149} \\
    5 & 0.9692 & 0.3247 & \textbf{0.8387} & 0.5214 & 0.5426 & 0.6054 \\
    \hline
  \end{tabular}\label{tab:comp}
  }
\end{table}

\begin{center}
\begin{tikzpicture}
\begin{axis}[
    title = Weighted F1-Score Comparison of proposed model with baseline,
	xtick = data,
	ylabel=Percentage,
	enlarge y limits  = 0.2,
    enlarge x limits  = 0.55,
	legend style={at={(0.5,-0.2)},
	anchor=north,legend columns=-1},
	ybar,
	width=\textwidth,
	height = .5\textwidth,
	x tick label style={/pgf/number format/1000 sep=},
	nodes near coords align={vertical},
	symbolic x coords = {Coarse-Grained, Fine-Grained},
]
\addplot coordinates {(Coarse-Grained, 83.98) (Fine-Grained, 49.54)};
\addplot coordinates {(Coarse-Grained, 84.11) (Fine-Grained, 52.01)};
\addplot coordinates {(Coarse-Grained, 79.79) (Fine-Grained, 22.40)};
\addplot coordinates {(Coarse-Grained, 83.45) (Fine-Grained, 45.94)};	
\addplot coordinates {(Coarse-Grained, 96.92) (Fine-Grained, 58.00)};
\addplot coordinates {(Coarse-Grained, 96.54) (Fine-Grained, 60.88)};
\addplot 
	[lightgray,fill]coordinates {(Coarse-Grained, 96.92) (Fine-Grained, 58.31)};
\addplot 
	[teal,fill] coordinates {(Coarse-Grained, 96.92) (Fine-Grained, 61.49)};
\addplot 
	[olive,fill] coordinates {(Coarse-Grained, 96.92) (Fine-Grained, 60.54)};
\legend{LR, SVM, RF, MLP, Sub.1, Sub.2, Sub.3, Sub.4, Sub.5}
\end{axis}
\end{tikzpicture}
\end{center}

\subsection{Analysis and Observations}
Table~\ref{tab5} includes the effect of \textit{Raw Representation} and \textit{Fine-tuned Representation} across coarse-grained and fine-grained tasks with XGBoost (XGB) classifier. It highlights the importance of \textit{Fine-tuned Representation}, as results improved with this. For fine-grained classification, the absolute f1-score improved by \textbf{0.24} by using fine-tuned representations. %One important point to note is that we should avoid overfitting while finetuning the m-BERT model. We overcome this problem by using a dropout layer before the classification. The weight decay in AdamW optimizer also helps in the regularization of model weights. Hence, our model generalizes well.
\begin{center}
    \begin{table}
    \centering
      \caption{Results with Raw and Fine-tune Representation with XGB on the validation dataset}
      \label{tab5}
      \begin{tabular}{|c| c| c|}
        \hline
        Representation & Coarse Grained F1-Score & Weighted Fine-grained F1-Score \\
        \hline
        Raw & 0.9298 & 0.3762 \\
        Fine-tune  &  \textbf{0.9692} & \textbf{0.6149} \\
        \hline
      \end{tabular}
    \end{table}
\end{center}

\section{Conclusion and Future Work}
In this paper, we presented a novel and simple ensemble based architecture for hostility detection in Hindi posts. We develop multiple architectures based on mBERT and different classifiers (i.e., ANN and XGBoost). Our proposed model outperformed all the existing baselines and secured 3rd rank in the competition.  In the future, we will try to tackle class imbalance using some weighting strategies during model training.

\bibliographystyle{splncs04}
\bibliography{mybibliography}

\begin{thebibliography}{8}
\bibitem{bhardwaj2020hostility} 
Mohit Bhardwaj, Md Shad Akhtar, Asif Ekbal, Amitava Das, and Tanmoy Chakraborty. 2020. Hostility Detection Dataset in Hindi.% \url{arXiv:2011.03588}

\bibitem{davidson2017automated} 
Thomas Davidson, Dana Warmsley, Michael Macy, and Ingmar Weber. 2017. Automated Hate Speech Detection and the Problem of Offensive Language.% \url{arXiv:1703.04009}

\bibitem{hossain-etal-2020-banfakenews}
Md Zobaer Hossain, Md Ashraful Rahman, Md Saiful Islam, and Sudipta Kar. 2020. BanFakeNews: A Dataset for Detecting Fake News in Bangla. In \textit{Proceedings of the 12th Language Resources and Evaluation Conference. European Language Resources Association, Marseille, France}, 2862–2871. \url{https://www.aclweb.org/anthology/2020.lrec-1.349/}
\bibitem{JHA20202324}
Vikas Kumar Jha, Hrudya P, Vinu P N, Vishnu Vijayan, and Prabaharan P. 2020. DHOT-Repository and Classification of Offensive Tweets in the Hindi Language. Procedia Computer Science 171 (2020), 2324 – 2333. \url{https://doi.org/10.1016/j.procs.2020.04.252} Third International Conference on Computing and Network Communications (CoCoNet’19).
\bibitem{leite2020toxic}
João A. Leite, Diego F. Silva, Kalina Bontcheva, and Carolina Scarton. 2020. Toxic Language Detection in Social Media for Brazilian Portuguese: New Dataset and Multilingual Analysis.
\bibitem{devlin2019bert}
Jacob Devlin and Ming-Wei Chang and Kenton Lee and Kristina Toutanova, 2019. BERT: Pre-training of Deep Bidirectional Transformers for Language Understanding. In \textit{Proceedings of the 2019 Conference of the North American Chapter of the Association for Computational Linguistics: Human Language Technologies, Volume 1 (Long and Short Papers)}, pages 4171–4186, Minneapolis, Minnesota. Association for Computational Linguistics

\bibitem{waseem-hovy-2016-hateful}
Zeerak Waseem and Dirk Hovy. 2016. Hateful Symbols or Hateful People? Predic-tive Features for Hate Speech Detection on Twitter. In \textit{Proceedings of the NAACLStudent Research Workshop. Association for Computational Linguistics, San Diego,California}, 88–93.  https://doi.org/10.18653/v1/N16-20132020-12-22 20:19. Page 4 of 1–4.
\bibitem{Chen_2016}
Chen, Tianqi, and Carlos Guestrin. “XGBoost.” \textit{Proceedings of the 22nd ACM SIGKDD International Conference on Knowledge Discovery and Data Mining (2016)}: n. pag. Crossref. Web.
\bibitem{safi_2020}
Safi Samghabadi, N.; Patwa, P.; PYKL, S.; Mukherjee, P.; Das, A.; and Solorio, T. 2020. Aggression and Misogyny Detection using BERT: A Multi-Task Approach. In \textit{Proceedings of the Second Workshop on Trolling, Aggression and Cyberbullying}, 126–131. Marseille, France: European Language Resources Association (ELRA).
% @inproceedings{patwa2021overview,
% title={Overview of CONSTRAINT 2021 Shared Tasks: Detecting English COVID-19 Fake News and Hindi Hostile Posts },
% author={Parth Patwa and Mohit Bhardwaj and Vineeth Guptha and Gitanjali Kumari and Shivam Sharma and Srinivas PYKL and Amitava Das and Asif Ekbal and Shad Akhtar and Tanmoy Chakraborty},
% booktitle = {Proceedings of the First Workshop on Combating Online Hostile Posts in Regional Languages during Emergency Situation ({CONSTRAINT})},
% year = {2021},
% publisher = {Springer},
% }
\bibitem{overview_paper}
Parth Patwa and Mohit Bhardwaj and Vineeth Guptha and Gitanjali Kumari and Shivam Sharma and Srinivas PYKL and Amitava Das and Asif Ekbal and Shad Akhtar and Tanmoy Chakraborty. 2021. Overview of CONSTRAINT 2021 Shared Tasks: Detecting English COVID-19 Fake News and Hindi Hostile Posts. In \textit{Proceedings of the First Workshop on Combating Online Hostile Posts in Regional Languages during Emergency Situation ({CONSTRAINT})}
% \bibitem{waseem2017understanding}
% Weerak Waseem, Thomas Davidson, Dana Warmsley, and Ingmar Weber. 2017.Understanding Abuse: A Typology of Abusive Language Detection Subtasks

\end{thebibliography}

\end{document}